\documentclass{article}
\usepackage{amsmath,graphicx,mlspconf}

\toappear{Preprint submitted to IEEE MLSP 2023}

\title{A Probabilistic Semi-Supervised Approach with Triplet Markov Chains}
\name{Katherine Morales, Yohan Petetin}
\address{Samovar, Telecom SudParis, Institut Polytechnique}

\usepackage{graphicx,xcolor,verbatim}
\usepackage{amsfonts}
\usepackage{multirow}
\usepackage{amsmath}
\usepackage{todonotes}
\usepackage[export]{adjustbox}
\usepackage{xcolor}
\DeclareMathOperator*{\argmax}{arg\,max}
\def\x{{\mathbf x}}
\def\z{{\mathbf z}}
\def\y{{\mathbf y}}
\def\v{{\mathbf v}}
\def\yl{{\mathbf y}_{T}^{\mathcal{L}}}
\def\yu{{\mathbf y}_{T}^{\mathcal{U}}}
\def\L{\mathcal{L}}
\def\U{\mathcal{U}}

\def\L{{\mathcal L}}

\def\p{p_{\theta}}
\def\q{q_\phi}
\def\Q{\tilde{Q}}

\begin{document}

\maketitle

\begin{abstract}
Triplet Markov chains are general generative models for sequential data which take into account three kinds of random variables: (noisy) observations, their associated discrete labels and latent variables which aim at strengthening the distribution of the observations and their associated labels.
However, in practice, we do not have at our disposal all the labels associated to the observations to estimate the parameters of such models. In this paper, we propose
a general framework based on a variational Bayesian inference to 
train parameterized triplet Markov chain models in a semi-supervised context. The generality of our approach enables us to derive semi-supervised
algorithms for a variety of generative models for sequential Bayesian classification.
\end{abstract}
\begin{keywords}
Generative Models; Variational Inference; Semi-Supervised Learning; Triplet Markov Chains.
\end{keywords}
\section{Introduction}
\label{sec:intro}
This paper focuses on semi-supervised 
learning for sequential Bayesian classification 
in general generative models. Let us start by
recalling the principle of Bayesian classification
and semi-supervised estimation.

\subsection{Sequential Bayesian classification}
We denote as $\x_T=(x_{0}, \dots, x_{T})$
a sequence of observed random variables (r.v.) and $\z_T=(z_{0}, \dots, z_{T})$ a sequence of latent r.v. We also introduce a sequence
of labels $\y_T=(y_{0}, \dots,\; y_{T})$ associated to the previous sequence $\x_T$. We will assume that  $x_t \in \mathbb{R}^{d_x}$, $z_t \in \mathbb{R}^{d_z}$ while the label
$y_t$ is discrete, so  $y_t \in \Omega=\{\omega_1,\dots,\omega_C\}$.
As far as notations are concerned, we do not distinguish r.v. and their realizations.
For example, $\x_T$ can represent a noisy grayscale image while $\y_T$ represents the original black and white image. For this application, the latent variable $z_t$ can be used to govern the conditional distribution of the noise given the original label.

When the labels associated to $\x_T$
are not observed, the objective associated to Bayesian classification consists in computing, for all $t$, the posterior distributions 
\begin{equation}
\label{eq:post_distrib}
p(y_t|\x_T)= 
\frac{\sum_{\y_{0:t-1},\y_{t+1:T}} \int p(\y_T,\x_T,\z_T) {\rm d}\z_T  } {\sum_{\y_T} \int p(\y_T,\x_T,\z_T) {\rm d}\z_T} \text{.}
\end{equation}
Consequently, we 
first need to define a parameterized model $\p(\x_T,\y_T,\z_T)$ which aims
at describing the r.v. involved in the
problem and from which 
it is possible to estimate 
$\theta$ and next to compute \eqref{eq:post_distrib} in a reasonable computational cost. 

The estimation of $\theta$ (i.e. the learning step) can be realized from sequences where we have at our disposal $(\x_T,\y_T)$ (supervised learning) or only $\x_T$ (unsupervised learning).
This general problem is commonly used in many fields, such as speech recognition~\cite{rabiner}, 
natural language processing~\cite{morwal2012named}, and activity recognition~\cite{reyes2016transition}.


\subsection{Semi-Supervised learning}
The problem we consider in this paper is a little bit different. In many real-world applications, it is expensive or impossible to obtain
labels for the entire sequence due to various reasons such as the high cost of labeling, the lack of expertise, or the lack of time. 
So from now on we assume that
we have at our disposal a sequence of observations
$\x_T$ with partially observed labels
and that i) we want to train relevant
generative models; and ii) we look for estimating the missing labels associated to each sequence. In other words, decomposing a sequence of labels $\y_T$ as
$$\y_T  = (\yl, \yu) \text{,}$$
where $\yl=\{y_t\}_{t\in \L}$ (resp. $\yu=\{y_t\}_{t\in \U}$) denotes the observed (resp. the unobserved) labels (so $\L$ (resp. $\U$) denotes the time index of observed (resp. unobserved) labels) we now look for estimating
$\theta$ from  $(\x_T,\yl)$, and next
computing, for all $t \in \U$,
$$p(y_t|\x_T,\yl)=\frac{\sum_{y_s \text{, }  s \in \U \backslash \{t\}} \int p(\y_T,\x_T,\z_T) {\rm d}\z_T  } { \sum_{y_s \text{, }  s \in \U} \int p(\y_T,\x_T,\z_T) {\rm d}\z_T} \text{.}$$

\subsection{Scope of the paper}
In this paper, we show that it is possible 
to propose a very general framework 
for semi-supervised learning in sequential generative 
models. In particular, we show that it is sufficient
to consider a generative model in which 
the triplet process $\{z_t,x_t,y_t\}_{t \geq 0}$ is Markovian.
As we will see later, such a general generative
model encompasses well known generative
models such that the Variational Sequential Labeler 
(VSL)~\cite{chen2019variational} or the Semi-supervised Variational Recurrent Neural Network (SVRNN)~\cite{butepage2019predicting}. Due to this general interpretation, 
we are able to propose new generative models that outperform
the previous ones for some semi-supervised learning tasks. The estimation 
of these general models is based on an adaptation of the
variational Bayesian framework~\cite{Jordan99anintroduction} for sequential and partially 
observed data.

The paper is organised as follows. In section \ref{sec:background}, we
recall the general triplet Markov chain (TMC) model \cite{wp-cras-chaines3,pieczynski2005triplet} and the principle of variational inference. In section \ref{sec:vi_tmc}, we show that our previous general models can be estimated with modified variational inference techniques in the context of semi-supervised learning. We next show that this general inference technique encompasses popular semi-supervised learning algorithms and we propose our Deep TMC 
model. Finally, in section \ref{sec:simulation}, we compare different approaches for the image segmentation problem. 






\section{Background}
\label{sec:background}
This section introduces the TMC model and the principle of
variational Bayesian inference techniques.

\subsection{Triplet Markov Chains}
Let us first start with the pair process $\{x_t,y_t\}_{t \geq 0}$. 
A popular model for describing this process is the hidden Markov chain model in which the sequences of label is assumed to be Markovian and the (noisy) observations are independent given the labels. In addition, the observation at a current time only depends 
on the label at the same time. In other words,
$$p(\x_T,\y_T)\overset{\rm HMC}{=}p(y_0)\prod_{t=1}^Tp(y_t|y_{t-1}) \prod_{t=0}^Tp(x_t|y_{t}) \text{.} $$
However, this model may be poor in practice due
to the Markovian assumption on the label. A simple way 
to relax it is to consider the pairwise Markov chain (PMC) model in which the pair
$\{x_t,y_t\}_{t \geq 0}$ is assumed to be Markovian,
$$p(\x_T,\y_T)\overset{\rm PMC}{=}p(y_0,x_0)\prod_{t=1}^Tp(y_t,x_t|y_{t-1},x_{t-1}) \text{.} $$
Even if this model is a direct generalization of the previous HMC model, it may be also unsatisfying in practice. The reason why is that it relies on the particular choice of the transition distribution $p(y_t,x_t|y_{t-1},x_{t-1})$ which may be difficult in practice. 
A simple way to address this problem is
to introduce a new latent process $\{z_t\}_{t \geq 0}$ which aims at making the previous distribution more robust. By denoting 
$ v_t=(z_t,x_t,y_t) \text{,}$ the general TMC model satisfies 
\begin{equation}
\label{eq:tmc}
p(\z_T,\x_T,\y_T)=p(\v_T)\overset{\rm TMC}{=}p(v_0)\prod_{t=1}^Tp(v_t|v_{t-1}) 
\end{equation}
and so, relies on the transition
distribution 
$$p(v_t|v_{t-1})=p(z_t,x_t,y_t|z_{t-1},x_{t-1},y_{t-1}) \text{.}$$
It encompasses the previous PMC and HMC models and also leads to a general class of generative models. 
For example, it is possible to first describe the distribution of the latent variable, then the one of the label given by the latent variable, and finally the one of the noisy observation given the latent variable and the label. In this case, the transition distribution reads as
$$p(v_t|v_{t-1})=p(z_t|v_{t-1})p(y_t|z_t,v_{t-1}) p(x_t|y_t,z_t,v_{t-1}) \text{.}$$
Even if the nature of the previous distributions is standard, the distribution
of interest
$$p(\x_T,\y_T)=\int p(\z_T,\x_T,\y_T) {\rm d}\z_T $$
can be complex.
In order to be used in practice,
we have to first choose a parameterized
transition distribution $\p(v_t|v_{t-1})$
and then estimate the parameter $\theta$ from partially labelled observations. 
This estimation step relies on the variational inference framework that we now recall.

\subsection{Variational Bayesian Inference}
\label{subsec:varinf}
In this section, we give up the temporal aspect of the problem and we only consider an observation $x$ and 
a latent r.v. $z$; the distribution
$\p(x,z)=p(z)p(x|z)$ is assumed to be known and we want to estimate
$\theta$ from a realization $x$.
A popular estimator is the Maximum-Likelihood (ML) estimate 
$\hat{\theta}=\argmax_{\theta} \p(x)$
due to its statistical properties \cite{White-MLE, Douc-ML-MIS}.
However, a direct maximization of $\p(x)$ is not always possible, particularly in
models with latent variables where the likelihood
$\p(x)=\int \p(x,z) {\rm d}z$ is not computable. 
In the variational inference framework, a variational lower bound called evidence lower bound 
(ELBO) on the log-likelihood  is optimized in order to estimate the parameters $\theta$~\cite{Jordan99anintroduction}.
This variational lower bound relies on the introduction of a parameterized 
variational distribution $\q(z|x)$ which aims at mimicking the true posterior $\p(z|x)$ and which is parameterized by a set of parameters $\phi$;
the ELBO reads
\begin{align}
\label{eq:elbo}
&\Q(\theta,\phi) = - \int \log \left(\frac{\q(z|x)}{\p(x,z)}\right) \q(z|x) {\rm d} z 
\end{align}
and satisfies, for all $(\theta,\phi)$
$$\log(\p(x)) \geq \Q(\theta,\phi) \text{.} $$
Equality holds if $\q(z|x)= \p(z|x)$.
In this particular case, 
the alternating  maximization w.r.t. $\theta$ and $\q$  of the ELBO, $\Q(\theta,\phi)$,
coincides with the EM algorithm \cite{variational-EM}. 

Variational inference consists in maximizing $Q(\theta,\phi)$ 
with respect to $(\theta,\phi)$ for a given class of distributions $\q$. 
The choice of the variational distribution $\q(z|x)$ is critical; 
$\q(z|x)$  should be close to $\p(z|x)$ but should also
be chosen in a such way that the associated ELBO can be exactly computed or easily approximated while remaining differentiable w.r.t. $(\theta,\phi)$. 
A simple way to approximate $\Q(\theta,\phi)$ with a Monte Carlo method  is to use the reparametrization 
trick~\cite{kingma2013auto} which consists in choosing a parametric
distribution $\q(z|x)$ such that a sample $z^{(i)} \sim q(z|x)$ can
be written as a differentiable function of $\phi$.


\section{Semi-supervised Variational Inference for TMCs}
\label{sec:vi_tmc}
Let us now turn back to the TMC model \eqref{eq:tmc} described by a parameterized transition
distribution
$\p(v_t|v_{t-1})$.

\subsection{Variational inference}

In order to estimate $\theta$ from 
partially labelled observations 
$(\x_T,\yl)$, we now want to maximize
the likelihood 
$$\p(\x_T, \yl)=\sum_{y_s \text{, }  s \in \U} \int \p(\z_T,\x_T,\y_T) {\rm d}\z_T$$
which is not computable in the general case.
We thus adapt the variational Bayesian
framework of section \ref{subsec:varinf},
where $x \leftarrow (\x_T, \yl)$ and
$z \leftarrow (\z_T, \yu)$. 

The ELBO \eqref{eq:elbo} now reads
\begin{align}
\label{eq:elbo_seq}
Q(\theta,\phi)&=  -  \sum_{y_s \text{, }  s \in \U} \int  \q(\z_T, \yu|\x_T, \yl) \times \nonumber \\  & \log \left(\frac{\q(\z_T,\yu  |\x_T,\yl)}{\p(\z_T,\x_T, \y_T)}\right)  {\rm d} \z_T \text{.}
\end{align}
\label{eq:TMC}
Let us now discuss on the 
computation of 
\eqref{eq:elbo_seq}.
First, it is worthwhile to remark 
that it does not depend on the choice of the generative model: 
any parameterized TMC model can be used since 
$ \p(\z_T,\x_T, \y_T)$ is known.
So its computation only depends
on the choice of the variational
distribution $\q(\z_T, \yu|\x_T, \yl)$. It can be factorized in two ways.
For sake of clarity, we will omit
the initial distribution of the variables at time $t=0$.
The first factorization 
coincides with 
\begin{align}
\label{eq:fact-1}
\q(\z_T, \yu|\x_T, \yl)=& \prod_{t=1}^T  \q(z_t|\z_{t-1},\y_{t-1},\x_T,\y_{t+1:T}^{\L}) \times \nonumber \\ & \prod_{t \in \U }^T 
\q(y_t|\y_{t-1},\z_t,\x_T,\y_{t+1:T}^{\L}) \text{,}
\end{align}
(remember that $\y_{t-1}=(\y_{t-1}^{\U},\y_{t-1}^{\L})$)
while the second one coincides
with 
\begin{align}
\label{eq:fact-2}
\q(\z_T, \yu|\x_T, \yl)=& \prod_{t=1}^T  \q(z_t|\z_{t-1},\y_{t},\x_T,\y_{t+1:T}^{\L}) \times \nonumber \\ & \! \! \prod_{t \in \U }^T 
\q(y_t|\y_{t-1},\z_{t-1},\x_T,\y_{t+1:T}^{\L}) \text{.}
\end{align}
As we explained in the previous section, the choice of the variational distribution depends on the generative model. In particular here, it has to take into account the factorization of the parameterized transition distribution $\p(v_t|v_{t-1})$.

It remains to compute the ELBO \eqref{eq:elbo_seq} which is nothing more than expectation according to $\q(\z_T, \yu|\x_T, \yl)$.
We thus propose to use a Monte-Carlo approximation based on the reparametrization trick in order
to obtain a differentiable approximation $\hat{Q}(\theta,\phi)$ of $Q(\theta,\phi)$. 
More precisely, we use the classical reparameterization trick to sample sequentially according to the continuous distribution $ q(z_t|\z_{t-1},\y_{t},\x_T,\y_{t+1:T}^{\L})$ (or 
 $q(z_t|\z_{t-1},\y_{t},\x_T,\y_{t+1:T}^{\L})$, while
 we use the Gumbel-Softmax (G-S) 
trick~\cite{maddison2016concrete, jang2016categorical} to sample according to $q(y_t|$ $\y_{t-1},\z_t,\x_T,\y_{t+1:T}^{\L})$ (or
$q(y_t|$ $\y_{t-1},\z_{t-1},\x_T,\y_{t+1:T}^{\L})$) since
the labels are discrete.

\subsection{Particular semi-supervised algorithms for TMCs}
\label{subsec:particular_cases}
As we have seen, semi-supervised algorithms for TMC depend on two key ingredients: the generative model described by the transition distribution $\p(v_t|v_{t-1})$ which
has an impact on the
performance of the model for a specific task (classification, prediction, detection, or generation),
and the choice of the variational distribution $\q(\z_T, \yu|\x_T, \yl)$. Actually, our general framework encompasses two existing solutions and we propose a new one.

\subsubsection{Variational Sequential Labeler (VSL)}
The VSL~\cite{chen2019variational} is a semi-supervised learning model for sequential data
and it has originally been proposed for the sequence labeling tasks in natural language processing.
It can be seen as a particular case of the general framework we derived in section \ref{sec:vi_tmc}.
This particular setting coincides with 
\begin{align}
\label{eq:vsl}
\p(v_t|v_{t-1}) \overset{\rm VSL}{=} \p(y_t|z_t) \p(z_t|x_{t-1}, z_{t-1}) \p(x_t|z_t) \text{,}
\end{align}
while the associated variational distribution
satisfies factorization \eqref{eq:fact-1}
with
\begin{align}
 \q(z_t|\z_{t-1},\y_{t-1},\x_T,\y_{t+1:T}^{\L})=\q(z_t|\x_T) \text{,} \\
 \q(y_t|\y_{t-1},\z_t,\x_T,\y_{t+1:T}^{\L})=\p(y_t|z_t) \text{.}
\end{align}
In this case, the ELBO \eqref{eq:elbo_seq} reduces to
\begin{align}
\label{eq:elbo_vsl}
Q(\theta,\phi) \overset{\rm VSL}{=}& \sum_{t \in \L} \int \q(z_{t}| \x_T) \log\p(y_{t}|z_{t}) {\rm d} z_t + \nonumber \\
&  \sum_{t=0}^T \int \q(z_{t}| \x_T) \Bigg[ \log \p(x_t|z_t) -\nonumber\\ 
& \quad \quad  \log \left(\frac{\q(z_t|\x_T)}{p(z_t|x_{t-1}, z_{t-1} )} \right)  \Bigg]  {\rm d} z_t \text{.}
\end{align}

It can be observed that it consists of two terms and that the previous assumptions enable to 
interpret it as an expectation according to $\q(\z_T|\x_T)$. Thus, it is not necessary to sample discrete variables according to the G-S trick. Moreover, a regularization term $\beta$ can be introduced in the second part of the ELBO in 
order to encourage good performance on labeled data while leveraging the context of the noisy image during reconstruction.
While this model simplifies the inference, it should be noted that in the generative process, the observation $x_t$ is conditionally independent of its associated label and may not
be adapted to some applications.


\subsubsection{ Semi-supervised Variational Recurrent Neural Network (SVRNN)}
\label{sec:svrnn}
The generative model used in the SVRNN can also be seen a particular version of the TMC model where the latent variable
$z_t$ consists of the pair $z_t=(z'_t, h_t)$. The associated
transition distribution reads:
\begin{align}
\label{eq:svrnn}
\!\!\!\! \p(v_t|v_{t-1}\!) \!=\!\p(y_t|v_{t-1}\!) \p(z_t|y_t,\! v_{t-1}\!) \p(x_t|y_t,z_t,\!v_{t-1}\!) \text{,}
\end{align}
where 
\begin{eqnarray*}
\p(y_t|v_{t-1}\!)&=& \p(y_t|h_{t-1})\text{,} \\
\p(z_t|y_t,v_{t-1})&=&\!\delta_{f_{\theta}(z'_t,y_t,x_t,h_{t-1})}(h_t) \!\!\times \!\p(z'_t|y_t, h_{t-1}) \text{,} \\
\p(x_t|y_t,z_t,\!v_{t-1}\!)&=& \p(x_t|y_t, z'_t, h_{t-1}) \text{,}
\end{eqnarray*}
and where $f_{\theta}$ is a deterministic 
function parameterized by a Recurrent Neural Network (RNN), for example.

On the other hand, the variational distribution $\q(\z_T, \yu|$ $ \x_T, \yl)$ satisfies
the factorization \eqref{eq:fact-2}
with
\begin{align*}
 q(z'_t|\z_{t-1},\y_{t},\x_T,\y_{t+1:T}^{\L})=  \q(z'_t| x_t, y_t, h_{t-1})\text{,} \\
q(y_t|\y_{t-1},\z_{t-1},\x_T,\y_{t+1:T}^{\L})= \q(y_t| x_t, h_{t-1}) \text{,}
\end{align*}
but their final ELBO does not
coincide with \eqref{eq:elbo_seq}. 
The reason why is that they derive it from the static case and add a penalization term that encourages $\p(y_t|v_{t-1})$ and $\q(y_t| x_t, h_{t-1})$ to be close to the empirical distribution of the data.



\subsubsection{Deep TMCs}
We finally present a very general TMC model from which one can apply any of the previous techniques.
The set of parameters $(\theta, \phi)$  
can be described by any differentiable flexible function $\psi(\cdot)$. 
In particular, we consider the case where the parameters are
produced by a (deep) neural network. 

Due to the different factorizations of the generating (resp. variational)
distributions, we consider a general notation
$\p(x_t | \cdot)$, $\p(z_t| \cdot)$ and $\p(y_t | \cdot)$ 
(resp. $\q(y_t | \cdot)$, $\q(z_t | \cdot)$) 
in order to avoid presenting a specific dependence between variables. 
These dependencies are specified for each model and are presented in the previous sections.

Let $\zeta(y_t; \cdot )$ and $ \varsigma(y_t;\cdot)$  (resp. $\lambda(x_t; \cdot )$; and $\eta(z_t;\cdot)$, $\tau(z_t;\cdot)$) 
be two probability distributions on $\Omega$
(resp. probability density functions on $\mathbb{R}^{d_x}$; and  $\mathbb{R}^{d_z}$ ).
The general model is described by:
\begin{align}
    \p(v_t|v_{t-1}) &=  \p(x_t|\cdot) \p(z_t|\cdot)\p(y_t|\cdot) \text{,} \nonumber\\
\label{eq:px}
    \p(x_t | \cdot) &= \lambda(x_t; \psi_{px}(\cdot) ) \text{,}\\
\label{eq:pz}
    \p(z_t | \cdot) &= \eta(z_t; \psi_{pz}(\cdot) ) \text{,}\\
\label{eq:py}
    \p(y_t | \cdot) &= \zeta(y_t;  \psi_{py}(\cdot)) \text{,}
\end{align}
(remember that $\psi_{px}(\cdot)$ denotes the parameters
of the distribution $\p(x_t | \cdot)$ and can depend 
on $v_{t-1}$ or $(v_{t-1},z_t,y_t)$ or $(v_{t-1},y_t)$, etc... according to the original factorization of $\p(v_t|v_{t-1})$). 
Finally, the variational distribution is given by
\begin{align}
\label{eq:qz}
    \q(z_t | \cdot) &=  \tau(z_t; \psi_{qz}(\cdot)) \text{,}\\
\label{eq:qy}
    \q(y_t | \cdot) &=  \varsigma(y_t;  \psi_{qy}(\cdot)) \text{.}
\end{align}
The parameters $\theta$ (resp. $\phi$) are derived from 
neural networks $(\psi_{px}, \psi_{pz}, \psi_{py})$ (resp.
$(\psi_{qz}, \psi_{qy})$).
Note that in the VLS model, $\psi_{qy}$ is not longer needed since 
the assumption $\q(y_t | z_t) = \p(y_t | z_t)$ is made.

\section{Simulations}
\label{sec:simulation}
In this section, we present the results of the proposed models on
semi-supervised binary image segmentation. Our goal is to recover the segmentation of a binary image 
($\Omega=\{\omega_1,\omega_2\}$) from the noisy observations
$\x_T$ when a partially segmentation $\yl$ is available.

In particular, $\zeta(y_t; \cdot )$ (resp. $\varsigma(y_t;\cdot)$) is set  as a Bernoulli distribution with parameters $\rho_{py,t}$ (resp. $\rho_{qy,t}$). 
As for the distribution  $\lambda(x_t; \cdot )$ (resp. $\eta(z_t;\cdot)$ and  $\tau(z_t;\cdot)$), we set it as a Gaussian distribution with parameters $[ \mu_{px,t}, {\rm diag}(\sigma_{px,t})]$ (resp.  $[ \mu_{pz,t}, {\rm diag}(\sigma_{pz,t})]$ and $[ \mu_{qz,t}, {\rm diag}(\sigma_{qz,t})]$).
where ${\rm diag(.)}$ denotes the diagonal matrix deduced from the values 
of $\sigma_{\cdot,t}$.

\subsection{Deep mTMC}

In our simulations, we consider three particular cases of the deep TMC model. We start with the deep minimal TMC (d-mTMC)~\cite{gangloff2023deep}, 
where the choice of parameters describes the transition:
\begin{align}
    \label{eq:mTMC}
    \p(v_t|v_{t-1}) \!\overset{\rm mTMC}{=} \!\p(y_t|y_{t-1}) \p(z_t|z_{t-1}) \p(x_t|y_t,z_t) \text{.}
\end{align}
So this model assumes a Markovian distribution 
for the labels and the latent variables aim at 
learning the distribution of the noise given the label and 
the latent variable.
In order to capture temporal dependencies in the input data and to have an efficient computation of the variational distribution for the d-mTMC model, we use a deterministic function to generate $\tilde{h}_t$ which  
takes as input $(x_t, y_t, z_t, \tilde{h}_{t-1})$. 
Then the variational distribution $\q(\z_T, \yu|$ $ \x_T, \yl)$ satisfies the factorization \eqref{eq:fact-2}
with $\q(z_t|x_t, y_t,\tilde{h}_{t-1} )$ and $\q(y_t|x_t, \tilde{h}_{t-1})$. 

In this case, the parameters are given by:
\begin{align*}
    &[\mu_{px,t}, \sigma_{px,t}]  = \psi_{px}(y_t, z_t),\\
    &[\mu_{pz,t}, \sigma_{pz,t}]  = \psi_{pz }(z_{t-1}), \\
    &\rho_{py,t}  = \psi_{px}(y_{t-1}),\\
     &[\mu_{qz,t}, \sigma_{qz,t}] = \psi_{qz}( x_t, y_t,\tilde{h}_{t-1})\text{, }\\
    &\rho_{qy,t}  = \psi_{qy}( x_t, \tilde{h}_{t-1}) \text{.}
\end{align*} 



\subsection{Experiments settings}
We used the Binary Shape Database~\cite{binaryimg} and focused 
on both \textit{cattle}-type and \textit{camel}-type images. 
To transform these images into a $1$-D signal ( $\x_T$ ), we used a Hilbert-Peano filling curve~\cite{sagan2012space}. 
To evaluate the models presented in Section~\ref{subsec:particular_cases},
we introduced non-linear blurring to highlight their ability to learn and
correct for signal corruption. 
More precisely, we generated an artificial noise for the \textit{cattle}-type by
generating $x_t$ according to
\begin{equation}
    \label{eq:noise_eq1}
    x_t| y_{t},x_{t-1} \sim \mathcal{N}\Big(\sin(a_{y_t}+x_{t-1});
    \sigma^2\Big),
\end{equation}
where $a_{\omega_1}=0$ , $a_{\omega_2} = 0.4$ and $\sigma^2=0.25$. 
We now consider the \textit{camel}-type image which is corrupted 
with a stationary multiplicative noise (non-elementary noise) given by
\begin{equation}
\label{eq:noise_eq2}
    x_t|y_t,z_t \sim\mathcal{N}\left(a_{y_t};b_{t_t}^2\right) * z_t,
\end{equation}
where $z_t\sim\mathcal{N}(0, 1)$, $a_{\omega_1}=0\text{, } a_{\omega_2} = 0.5$ and $b_{\omega_1}=b_{\omega_2}=0.2$. 

\begin{figure}[htb]
    \begin{minipage}[b]{0.30\linewidth}
      \centering
      \centerline{\includegraphics[width=\textwidth, cfbox=black 1pt 0pt]{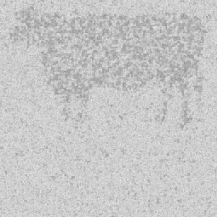}}
      \centerline{(a) $\x_T$}\medskip
    \end{minipage}
    \hfill
    \begin{minipage}[b]{.30\linewidth}
      \centering
      \centerline{\includegraphics[width=\textwidth, cfbox=black 1pt 0pt]{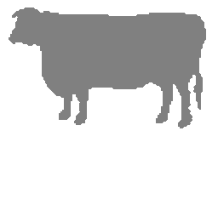}}
      \centerline{(b) True labels}\medskip
    \end{minipage}
    \hfill
    \begin{minipage}[b]{0.30\linewidth}
      \centering
      \centerline{\includegraphics[width=\textwidth,cfbox=black 1pt 0pt]{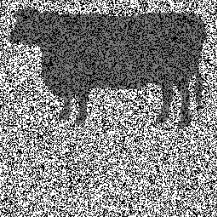}}
      \centerline{(c) $\U$ -  40\%  }\medskip
    \end{minipage}
    \\
    \begin{minipage}[b]{0.30\linewidth}
        \centering
        \centerline{\includegraphics[width=\textwidth,cfbox=black 1pt 0pt]{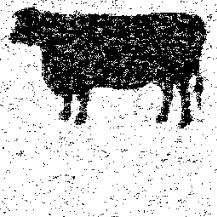}}
        \centerline{(d) VLS}\medskip      \end{minipage}
      \hfill
      \begin{minipage}[b]{.30\linewidth}
        \centering
        \centerline{\includegraphics[width=\textwidth,cfbox=black 1pt 0pt]{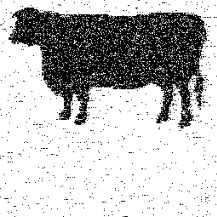}}
        \centerline{(e) SVRNN}\medskip
      \end{minipage}
      \hfill
      \begin{minipage}[b]{0.30\linewidth}
        \centering
        \centerline{\includegraphics[width=\textwidth,cfbox=black 1pt 0pt]{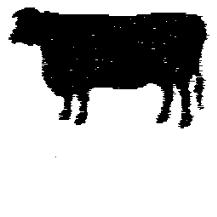}}
        \centerline{(f) d-mTMC}\medskip
      \end{minipage}
    \caption{Semi-supervised image segmentation with d-TMC models with $40\%$ of unlabeled pixels.}
    \label{fig:res_cow40}
\end{figure}

The generated images are presented in Fig.\ref{fig:res_cow40}(a) 
and Fig.\ref{fig:res_camel60}(a), respectively. 
More details about the image generation process are
available in~\cite{gangloff2023deep}. Additionally, we randomly selected
pixels $y_t \in \yl$, with a percentage of the pixels being labeled and the rest considered unobserved (\textit{e.g.} Fig.\ref{fig:res_cow40}(c) 
and Fig.\ref{fig:res_camel60}(c) ). 

Each model was trained using stochastic gradient descent to optimize the negative associated ELBO, with the Adam optimizer~\cite{adam}. The neural networks $\psi_{(\cdot)}$'s were designed with two hidden layers using rectified linear units and appropriate outputs, such as linear, softplus, and sigmoid. To ensure a fair comparison, we matched the total number of parameters of all models to be approximately equal. As a result, the number of hidden units for each hidden layer differs for each model. In fact, the SVRNN, mTMC, and VLS models have 22, 25, and 41 hidden units, respectively.
We used an RNN cell to generate $\tilde{h}_t$ (resp. $h_t$) for the d-mTMC (resp. SVRNN) model. In the VLS model, we used the parameterization approach for $\q(z_t|\x_T)$ presented in~\cite{chen2019variational}, which involves using a bi-directional RNN cell. We also set the regularization term to $0,1$.

\subsection{Results}
\label{subsec:results}
The performance of the models is evaluated in terms of
the error rate (ER) of the reconstruction of the unobserved pixels.  
Table~\ref{tab:error_rates} presents the error rates obtained for reconstructing unobserved pixels on different images. The notation \textit{image $\%$} is used to indicate the specific image and the percentage of unobserved labels in the image. 
As shown in the table, the d-mTMC consistently outperforms the VSL and the SVRNN, achieving a lower error rate for each case. 

Additionally, we observe that when dealing with elementary noise, the performance of the VLS model is superior to that of SVRNN. However, this capability is lost as we increase the percentage of unobserved labels, even with elementary noise.
\begin{table}[htb]
\centering
\begin{tabular}{l|r|r|l|}
\cline{2-4}
                            & \multicolumn{1}{l|}{\textit{Cattle} $40\%$} & \multicolumn{1}{l|}{\textit{Camel} $40\%$} & \textit{Camel} $60\%$                      \\ \hline
\multicolumn{1}{|l|}{VSL}   & 15,64\%                    & 41,84\%                      & \multicolumn{1}{r|}{41,80\%} \\ \hline
\multicolumn{1}{|l|}{SVRNN} & 16,55\%                    & 12,12\%                      & \multicolumn{1}{r|}{  21,38\%}                            \\ \hline
\multicolumn{1}{|l|}{d-mTMM}   & \textbf{1,93}\%                     &\textbf{ 2,60}\%                       & \multicolumn{1}{r|}{\textbf{3,62}\%}  \\ \hline
\end{tabular}
\caption{Error rates of the reconstruction of the unobserved pixels on different images with different percentages of unobserved pixels.}
\label{tab:error_rates}
\end{table}

\begin{figure}[htb]
    \begin{minipage}[b]{0.30\linewidth}
      \centering
      \centerline{\includegraphics[width=\textwidth, cfbox=black 1pt 0pt]{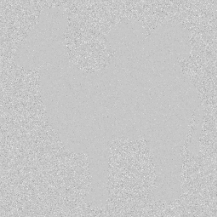}}
      \centerline{(a) $\x_T$}\medskip
    \end{minipage}
    \hfill
    \begin{minipage}[b]{.30\linewidth}
      \centering
      \centerline{\includegraphics[width=\textwidth, cfbox=black 1pt 0pt]{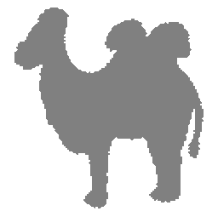}}
      \centerline{(b) True labels}\medskip
    \end{minipage}
    \hfill
    \begin{minipage}[b]{0.30\linewidth}
      \centering
      \centerline{\includegraphics[width=\textwidth,cfbox=black 1pt 0pt]{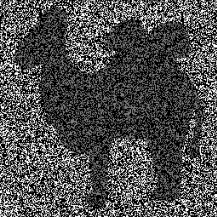}}
      \centerline{(c) $\U$ - 60\%  }\medskip
    \end{minipage}
    \\
    \begin{minipage}[b]{0.30\linewidth}
        \centering
        \centerline{\includegraphics[width=\textwidth,cfbox=black 1pt 0pt]{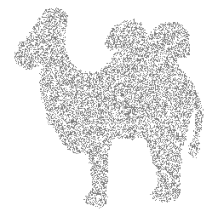}}
        \centerline{(d) VLS}\medskip      \end{minipage}
      \hfill
      \begin{minipage}[b]{.30\linewidth}
        \centering
        \centerline{\includegraphics[width=\textwidth,cfbox=black 1pt 0pt]{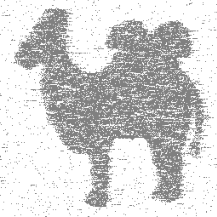}}
        \centerline{(e) SVRNN}\medskip
      \end{minipage}
      \hfill
      \begin{minipage}[b]{0.30\linewidth}
        \centering
        \centerline{\includegraphics[width=\textwidth,cfbox=black 1pt 0pt]{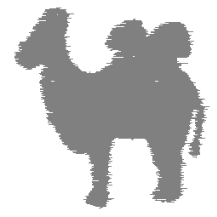}}
        \centerline{(f) d-mTMC}\medskip
      \end{minipage}
    \caption{Semi-supervised image segmentation with d-TMC models with $60\%$ of unlabeled pixels.}
    \label{fig:res_camel60}
\end{figure}

Moreover, our algorithm achieves superior performance with a more complex noise (the \textit{camel}-type image).
Fig. \ref{fig:res_camel60} displays the performance of our proposed algorithm compared to the VSL and the SVRNN on the \textit{camel}-type image with $60\%$ of unobserved labels.

\section{Conclusion}
\label{sec:conclusion}
In this paper, we have presented a general 
framework for semi-supervised estimation in generative
models. In particular, by considering the TMC model, we have shown that it is possible
to obtain a wide variety of generative models and
to estimate them in the common framework of
variational inference in the case where only
a part of the observations are labelled.
Our experiments demonstrate the effectiveness of the proposed approach in achieving state-of-the-art performance on the task of image segmentation.


\bibliographystyle{IEEEbib}
\bibliography{refs}

\begin{thebibliography}{10}

\bibitem{rabiner}
L.~R. Rabiner,
\newblock ``A tutorial on {H}idden {M}arkov {M}odels and selected applications
  in speech recognition,''
\newblock {\em Proceedings of the IEEE}, vol. 77, no. 2, pp. 257--286, February
  1989.

\bibitem{morwal2012named}
S.~Morwal, N.~Jahan, and D.~Chopra,
\newblock ``Named entity recognition using hidden markov model (hmm),''
\newblock {\em International Journal on Natural Language Computing (IJNLC)
  Vol}, vol. 1, 2012.

\bibitem{reyes2016transition}
J.-L. Reyes-Ortiz, L.~Oneto, A.~Sam\`{a}, X.~Parra, and D.~Anguita,
\newblock ``Transition-aware human activity recognition using smartphones,''
\newblock {\em Neurocomputing}, vol. 171, pp. 754–767, 2016.

\bibitem{chen2019variational}
M.~Chen, Q.~Tang, K.~Livescu, and K.~Gimpel,
\newblock ``Variational sequential labelers for semi-supervised learning,''
\newblock {\em arXiv preprint arXiv:1906.09535}, 2019.

\bibitem{butepage2019predicting}
J.~Butepage, H.~Kjellstrom, and D.~Kragic,
\newblock ``Predicting the what and how-a probabilistic semi-supervised
  approach to multi-task human activity modeling,''
\newblock in {\em Proceedings of the IEEE/CVF Conference on Computer Vision and
  Pattern Recognition Workshops}, 2019, pp. 0--0.

\bibitem{Jordan99anintroduction}
I.~J. Michael, G.~Zoubin, T.~S. Jaakola, and L.~K. Saul,
\newblock ``An introduction to variational methods for graphical models,''
\newblock in {\em MACHINE LEARNING}. 1999, pp. 183--233, MIT Press.

\bibitem{wp-cras-chaines3}
W.~Pieczynski,
\newblock ``Chaines de {M}arkov triplet,''
\newblock {\em Comptes Rendus de l'Academie des Sciences - Mathematiques}, vol.
  335, pp. 275--278, 2002,
\newblock in French.

\bibitem{pieczynski2005triplet}
W.~Pieczynski and F.~Desbouvries,
\newblock ``On triplet {M}arkov chains,''
\newblock in {\em International Symposium on Applied Stochastic Models and Data
  Analysis, ({ASMDA})}, 2005.

\bibitem{White-MLE}
H.~White,
\newblock ``Maximum likelihood estimation of misspecified models,''
\newblock {\em Econometrica}, vol. 50, no. 1, pp. 1--25, January 1982.

\bibitem{Douc-ML-MIS}
R.~Douc and E.~Moulines,
\newblock ``Asymptotic properties of the maximum likelihood estimation in
  misspecified hidden {M}arkov models,''
\newblock {\em Annals of Statistics}, vol. 40, no. 5, pp. 2697--2732, 2012.

\bibitem{variational-EM}
D.~G. Tzikas, A.~C. Likas, and N.~P. Galatsanos,
\newblock ``The variational approximation for {B}ayesian inference,''
\newblock {\em IEEE Signal Processing Magazine}, vol. 25, no. 6, pp. 131--146,
  2008.

\bibitem{kingma2013auto}
D.~P. Kingma and M.~Welling,
\newblock ``Auto-encoding variational {B}ayes,''
\newblock in {\em 2nd International Conference on Learning Representations,
  {ICLR}}, 2014.

\bibitem{maddison2016concrete}
C~J. Maddison, A.~Mnih, and Y.~W. Teh,
\newblock ``The concrete distribution: A continuous relaxation of discrete
  random variables,''
\newblock {\em arXiv preprint arXiv:1611.00712}, 2016.

\bibitem{jang2016categorical}
E.~Jang, S.~Gu, and B.~Poole,
\newblock ``Categorical reparameterization with gumbel-softmax,''
\newblock {\em arXiv preprint arXiv:1611.01144}, 2016.

\bibitem{gangloff2023deep}
H.~Gangloff, K.~Morales, and Y.~Petetin,
\newblock ``Deep parameterizations of pairwise and triplet markov models for
  unsupervised classification of sequential data,''
\newblock {\em Computational Statistics \& Data Analysis}, vol. 180, pp.
  107663, 2023.

\bibitem{binaryimg}
LEMS-Computer~Vision Group,
\newblock ``Binary shape,''
\newblock {\em
  https://vision.lems.brown.edu/content/available-software-and-databases}.

\bibitem{sagan2012space}
H.~Sagan,
\newblock {\em Space-filling curves},
\newblock Springer, 2012.

\bibitem{adam}
D.~Kingma and J.~Ba,
\newblock ``Adam: A method for stochastic optimization,''
\newblock {\em International conference on learning representations}, 12 2014.

\end{thebibliography}

\end{document}